# Improving Lexical Choice in Neural Machine Translation

**Toan Q. Nguyen** and **David Chiang**
Department of Computer Science and Engineeering
University of Notre Dame
{tnguye28,dchiang}@nd.edu

## Abstract

We explore two solutions to the problem of mistranslating rare words in neural machine translation. First, we argue that the standard output layer, which computes the inner product of a vector representing the context with all possible output word embeddings, rewards frequent words disproportionately, and we propose to fix the norms of both vectors to a constant value. Second, we integrate a simple lexical module which is jointly trained with the rest of the model. We evaluate our approaches on eight language pairs with data sizes ranging from 100k to 8M words, and achieve improvements of up to +4.3 BLEU, surpassing phrase-based translation in nearly all settings.

## 1 Introduction

Neural network approaches to machine translation (Sutskever et al., 2014; Bahdanau et al., 2015; Luong et al., 2015a; Gehring et al., 2017) are appealing for their single-model, end-to-end training process, and have demonstrated competitive performance compared to earlier statistical approaches (Koehn et al., 2007; Junczys-Dowmunt et al., 2016). However, there are still many open problems in NMT (Koehn and Knowles, 2017). One particular issue is mistranslation of rare words. For example, consider the Uzbek sentence:

*Source:* Ammo muammolar hali ko'p, deydi amerikalik olim Entoni Fauchi.

*Reference:* But still there are many problems, says American scientist Anthony Fauci.

*Baseline NMT:* But there is still a lot of problems, says James Chan.

At the position where the output should be *Fauci*, the NMT model's top three candidates are *Chan*, *Fauci*, and *Jenner*. All three surnames occur in the training data with reference to immunologists: Fauci is the director of the National Institute of Allergy and Infectious Diseases, Margaret (not James) Chan is the former director of the World Health Organization, and Edward Jenner invented smallpox vaccine. But *Chan* is more frequent in the training data than *Fauci*, and *James* is more frequent than either *Anthony* or *Margaret*.

Because NMT learns word representations in continuous space, it tends to translate words that "seem natural in the context, but do not reflect the content of the source sentence" (Arthur et al., 2016). This coincides with other observations that NMT's translations are often fluent but lack accuracy (Wang et al., 2017b; Wu et al., 2016).

Why does this happen? At each time step, the model's distribution over output words $e$ is

$$p(e) \propto \exp\left(W_e \cdot \tilde{h} + b_e\right)$$

where $W_e$ and $b_e$ are a vector and a scalar depending only on $e$, and $\tilde{h}$ is a vector depending only on the source sentence and previous output words. We propose two modifications to this layer. First, we argue that the term $W_e \cdot \tilde{h}$, which measures how well $e$ fits into the context $\tilde{h}$, favors common words disproportionately, and show that it helps to fix the norm of both vectors to a constant. Second, we add a new term representing a more direct connection from the source sentence, which allows the model to better memorize translations of rare words.

Below, we describe our models in more detail. Then we evaluate our approaches on eight language pairs, with training data sizes ranging from 100k words to 8M words, and show improvements of up to +4.3 BLEU, surpassing phrase-based translation in nearly all settings. Finally, we provide some analysis to better understand why our modifications work well.

## 2 Neural Machine Translation

Given a source sequence $f = f_1 f_2 \cdots f_m$, the goal of NMT is to find the target sequence $e =$

|  | ha-en | tu-en | hu-en |
|---|---|---|---|
| untied embeddings | 17.2 | 11.5 | 26.5 |
| tied embeddings | 17.4 | 13.8 | 26.5 |
| don't normalize $\tilde{h}_t$ | 18.6 | 14.2 | 27.1 |
| normalize $\tilde{h}_t$ | 20.5 | 16.1 | 28.8 |

Table 1: Preliminary experiments show that tying target embeddings with output layer weights performs as well as or better than the baseline, and that normalizing $\tilde{h}$ is better than not normalizing $\tilde{h}$. All numbers are BLEU scores on development sets, scored against tokenized references.

$e_1 e_2 \cdots e_n$ that maximizes the objective function:

$$\log p(e \mid f) = \sum_{t=1}^{n} \log p(e_t \mid e_{<t}, f).$$

We use the *global attentional* model with *general scoring function* and *input feeding* by Luong et al. (2015a). We provide only a very brief overview of this model here. It has an encoder, an attention, and a decoder. The encoder converts the words of the source sentence into *word embeddings*, then into a sequence of *hidden states*. The decoder generates the target sentence word by word with the help of the attention. At each time step $t$, the attention calculates a set of *attention weights* $a_t(s)$. These attention weights are used to form a weighted average of the encoder hidden states to form a *context vector* $c_t$. From $c_t$ and the hidden state of the decoder are computed the *attentional hidden state* $\tilde{h}_t$. Finally, the predicted probability distribution of the $t$'th target word is:

$$p(e_t \mid e_{<t}, f) = \text{softmax}(W^o \tilde{h}_t + b^o). \quad (1)$$

The rows of the output layer's weight matrix $W^o$ can be thought of as embeddings of the output vocabulary, and sometimes are in fact tied to the embeddings in the input layer, reducing model size while often achieving similar performance (Inan et al., 2017; Press and Wolf, 2017). We verified this claim on some language pairs and found out that this approach usually performs better than without tying, as seen in Table 1. For this reason, we always tie the target embeddings and $W^o$ in all of our models.

## 3 Normalization

The output word distribution (1) can be written as:

$$p(e) \propto \exp\left(\|W_e\| \|\tilde{h}\| \cos \theta_{W_e, \tilde{h}} + b_e\right),$$

where $W_e$ is the embedding of $e$, $b_e$ is the $e$'th component of the bias $b^o$, and $\theta_{W_e, \tilde{h}}$ is the angle between $W_e$ and $\tilde{h}$. We can intuitively interpret the terms as follows. The term $\|\tilde{h}\|$ has the effect of sharpening or flattening the distribution, reflecting whether the model is more or less certain in a particular context. The cosine similarity $\cos \theta_{W_e, \tilde{h}}$ measures how well $e$ fits into the context. The bias $b_e$ controls how much the word $e$ is generated; it is analogous to the language model in a log-linear translation model (Och and Ney, 2002).

Finally, $\|W_e\|$ also controls how much $e$ is generated. Figure 1 shows that it generally correlates with frequency. But because it is multiplied by $\cos \theta_{W_e, \tilde{h}}$, it has a stronger effect on words whose embeddings have direction similar to $\tilde{h}$, and less effect or even a negative effect on words in other directions. We hypothesize that the result is that the model learns $\|W_e\|$ that are disproportionately large.

For example, returning to the example from Section 1, these terms are:

| $e$ | $\|W_e\|$ | $\|\tilde{h}\|$ | $\cos \theta_{W_e, \tilde{h}}$ | $b_e$ | logit |
|---|---|---|---|---|---|
| Chan | 5.25 | 19.5 | 0.144 | $-1.53$ | 13.2 |
| Fauci | 4.69 | 19.5 | 0.154 | $-1.35$ | 12.8 |
| Jenner | 5.23 | 19.5 | 0.120 | $-1.59$ | 10.7 |

Observe that $\cos \theta_{W_e, \tilde{h}}$ and even $b_e$ both favor the correct output word *Fauci*, whereas $\|W_e\|$ favors the more frequent, but incorrect, word *Chan*. The most frequently-mentioned immunologist trumps other immunologists.

To solve this issue, we propose to fix the norm of all target word embeddings to some value $r$. Following the weight normalization approach of Salimans and Kingma (2016), we reparameterize $W_e$ as $r \frac{v_e}{\|v_e\|}$, but keep $r$ fixed.

A similar argument could be made for $\|\tilde{h}_t\|$: because a large $\|\tilde{h}_t\|$ sharpens the distribution, causing frequent words to more strongly dominate rare words, we might want to limit it as well. We compared both approaches on a development set and found that replacing $\tilde{h}_t$ in equation (1) with $r \frac{\tilde{h}_t}{\|\tilde{h}_t\|}$ indeed performs better, as shown in Table 1.

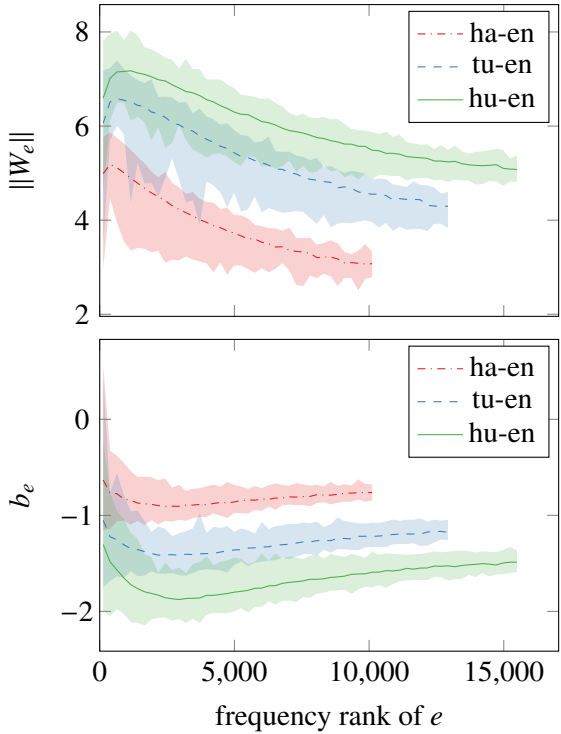

Figure 1: The word embedding norm $\|W_e\|$ generally correlates with the frequency of $e$, except for the most frequent words. The bias $b_e$ has the opposite behavior. The plots show the median and range of bins of size 256.

| | tokens $\times 10^6$ | vocab $\times 10^3$ | layers num/size |
|---|---|---|---|
| ta-en | 0.2/0.1 | 4.0/3.4 | 1/512 |
| ur-en | 0.2/0.2 | 4.2/4.2 | 1/512 |
| ha-en | 0.8/0.8 | 10.6/10.4 | 2/512 |
| tu-en | 0.8/1.1 | 21.1/13.3 | 2/512 |
| uz-en | 1.5/1.9 | 29.8/17.4 | 2/512 |
| hu-en | 2.0/2.3 | 27.3/15.7 | 2/512 |
| en-vi | 2.1/2.6 | 17.0/7.7 | 2/512 |
| en-ja (BTEC) | 3.6/5.0 | 17.8/21.8 | 4/768 |
| en-ja (KFTT) | 7.8/8.0 | 48.2/49.1 | 4/768 |

Table 2: Statistics of data and models: effective number of training source/target tokens, source/target vocabulary sizes, number of hidden layers and number of units per layer.

## 4 Lexical Translation

The attentional hidden state $\tilde{h}$ contains information not only about the source word(s) corresponding to the current target word, but also the contexts of those source words and the preceding context of the target word. This could make the model prone to generate a target word that fits the context but doesn't necessarily correspond to the source word(s). Count-based statistical models, by contrast, don't have this problem, because they simply don't model any of this context. Arthur et al. (2016) try to alleviate this issue by integrating a count-based lexicon into an NMT system. However, this lexicon must be trained separately using GIZA++ (Och and Ney, 2003), and its parameters form a large, sparse array, which can be difficult to store in GPU memory.

We propose instead to use a simple feedforward neural network (FFNN) that is trained jointly with the rest of the NMT model to generate a target word based directly on the source word(s). Let $f_s$ ($s = 1, \ldots, m$) be the embeddings of the source words. We use the attention weights to form a weighted average of the embeddings (not the hidden states, as in the main model) to give an average source-word embedding at each decoding time step $t$:

$$f_t^\ell = \tanh \sum_s a_t(s) f_s.$$

Then we use a one-hidden-layer FFNN with skip connections (He et al., 2016):

$$h_t^\ell = \tanh(W f_t^\ell) + f_t^\ell$$

and combine its output with the decoder output to get the predictive distribution over output words at time step $t$:

$$p(y_t \mid y_{<t}, x) = \text{softmax}(W^o \tilde{h}_t + b^o + W^\ell h_t^\ell + b^\ell).$$

For the same reasons that were given in Section 3 for normalizing $\tilde{h}_t$ and the rows of $W_t^o$, we normalize $h_t^\ell$ and the rows of $W^\ell$ as well. Note, however, that we do not tie the rows of $W^\ell$ with the word embeddings; in preliminary experiments, we found this to yield worse results.

## 5 Experiments

We conducted experiments testing our normalization approach and our lexical model on eight language pairs using training data sets of various sizes. This section describes the systems tested and our results.

### 5.1 Data

We evaluated our approaches on various language pairs and datasets:

- Tamil (ta), Urdu (ur), Hausa (ha), Turkish (tu), and Hungarian (hu) to English (en), using data from the LORELEI program.

- English to Vietnamese (vi), using data from the IWSLT 2015 shared task.[1]

- To compare our approach with that of Arthur et al. (2016), we also ran on their English to Japanese (ja) KFTT and BTEC datasets.[2]

We tokenized the LORELEI datasets using the default Moses tokenizer, except for Urdu-English, where the Urdu side happened to be tokenized using Morfessor FlatCat ($w = 0.5$). We used the preprocessed English-Vietnamese and English-Japanese datasets as distributed by Luong et al., and Arthur et al., respectively. Statistics about our data sets are shown in Table 2.

### 5.2 Systems

We compared our approaches against two baseline NMT systems:

**untied**, which does not tie the rows of $W_o$ to the target word embeddings, and

**tied**, which does.

In addition, we compared against two other baseline systems:

**Moses**: The Moses phrase-based translation system (Koehn et al., 2007), trained on the same data as the NMT systems, with the same maximum sentence length of 50. No additional data was used for training the language model. Unlike the NMT systems, Moses used the full vocabulary from the training data; unknown words were copied to the target sentence.

**Arthur:** Our reimplementation of the discrete lexicon approach of Arthur et al. (2016). We only tried their `auto` lexicon, using GIZA++ (Och and Ney, 2003), integrated using their `bias` approach. Note that we also tied embedding as we found it also helped in this case.

Against these baselines, we compared our new systems:

**fixnorm**: The normalization approach described in Section 3.

**fixnorm+lex**: The same, with the addition of the lexical translation module from Section 4.

---

[1] https://nlp.stanford.edu/projects/nmt/
[2] http://isw3.naist.jp/~philip-a/emnlp2016/

### 5.3 Details

**Model** For all NMT systems, we fed the source sentences to the encoder in reverse order during both training and testing, following Luong et al. (2015a). Information about the number and size of hidden layers is shown in Table 2. The word embedding size is always equal to the hidden layer size.

Following common practice, we only trained on sentences of 50 tokens or less. We limited the vocabulary to word types that appear no less than 5 times in the training data and map the rest to UNK. For the English-Japanese and English-Vietnamese datasets, we used the vocabulary sizes reported in their respective papers (Arthur et al., 2016; Luong and Manning, 2015).

For **fixnorm**, we tried $r \in \{3, 5, 7\}$ and selected the best value based on the development set performance, which was $r = 5$ except for English-Japanese (BTEC), where $r = 7$. For **fixnorm+lex**, because $W_s \tilde{h}_t + W^\ell h_t^\ell$ takes on values in $[-2r^2, 2r^2]$, we reduced our candidate $r$ values by roughly a factor of $\sqrt{2}$, to $r \in \{2, 3.5, 5\}$. A radius $r = 3.5$ seemed to work the best for all language pairs.

**Training** We trained all NMT systems with Adadelta (Zeiler, 2012). All parameters were initialized uniformly from $[-0.01, 0.01]$. When a gradient's norm exceeded 5, we normalized it to 5. We also used dropout on non-recurrent connections only (Zaremba et al., 2014), with probability 0.2. We used minibatches of size 32. We trained for 50 epochs, validating on the development set after every epoch, except on English-Japanese, where we validated twice per epoch. We kept the best checkpoint according to its BLEU on the development set.

**Inference** We used beam search with a beam size of 12 for translating both the development and test sets. Since NMT often favors short translations (Cho et al., 2014), we followed Wu et al. (2016) in using a modified score $s(e \mid f)$ in place of log-probability:

$$s(e \mid f) = \frac{\log p(e \mid f)}{lp(e)}$$
$$lp(e) = \frac{(5 + |e|)^\alpha}{(5 + 1)^\alpha}$$

We set $\alpha = 0.8$ for all of our experiments.

Finally, we applied a postprocessing step to replace each UNK in the target translation with the source word with the highest attention score (Luong et al., 2015b).

**Evaluation** For translation into English, we report case-sensitive NIST BLEU against detokenized references. For English-Japanese and English-Vietnamese, we report tokenized, case-sensitive BLEU following Arthur et al. (2016) and Luong and Manning (2015). We measure statistical significance using bootstrap resampling (Koehn, 2004).

## 6 Results and Analysis

### 6.1 Overall

Our results are shown in Table 3. First, we observe, as has often been noted in the literature, that NMT tends to perform poorer than PBMT on low resource settings (note that the rows of this table are sorted by training data size).

Our **fixnorm** system alone shows large improvements (shown in parentheses) relative to **tied**. Integrating the lexical module (**fixnorm+lex**) adds in further gains. Our **fixnorm+lex** models surpass Moses on all tasks except Urdu- and Hausa-English, where it is 1.6 and 0.7 BLEU short respectively.

The method of Arthur et al. (2016) does improve over the baseline NMT on most language pairs, but not by as much and as consistently as our models, and often not as well as Moses. Unfortunately, we could not replicate their approach for English-Japanese (KFTT) because the lexical table was too large to fit into the computational graph.

For English-Japanese (BTEC), we note that, due to the small size of the test set, all systems except for Moses are in fact not significantly different from **tied** ($p > 0.01$). On all other tasks, however, our systems significantly improve over **tied** ($p < 0.01$).

### 6.2 Impact on translation

In Table 4, we show examples of typical translation mistakes made by the baseline NMT systems. In the Uzbek example (top), **untied** and **tied** have confused *34* with *UNK* and *700*, while in the Turkish one (middle), they incorrectly output other proper names, *Afghan* and *Myanmar*, for the proper name *Kenya*. Our systems, on the other hand, translate these words correctly.

The bottom example is the one introduced in Section 1. We can see that our **fixnorm** approach does not completely solve the mistranslation issue, since it translates *Entoni Fauchi* to *UNK UNK* (which is arguably better than *James Chan*). On the other hand, **fixnorm+lex** gets this right. To better understand how the lexical module helps in this case, we look at the top five translations for the word *Fauci* in **fixnorm+lex**:

| $e$ | $\cos\theta_{W_e,\tilde{h}}$ | $\cos\theta_{W_e^l,h_l}$ | $b_e + b_e^l$ | logit |
|---|---|---|---|---|
| Fauci | 0.522 | 0.762 | −8.71 | 7.0 |
| UNK | 0.566 | −0.009 | −1.25 | 5.6 |
| Anthony | 0.263 | 0.644 | −8.70 | 2.4 |
| Ahmedova | 0.555 | 0.173 | −8.66 | 0.3 |
| Chan | 0.546 | 0.150 | −8.73 | −0.2 |

As we can see, while $\cos\theta_{W_e,\tilde{h}}$ might still be confused between similar words, $\cos\theta_{W_e^l,h_l}$ significantly favors *Fauci*.

### 6.3 Alignment and unknown words

Both our baseline NMT and **fixnorm** models suffer from the problem of shifted alignments noted by Koehn and Knowles (2017). As seen in Figure 2a and 2b, the alignments for those two systems seem to shift by one word to the left (on the source side). For example, *nói* should be aligned to *said* instead of *Telekom*, and so on. Although this is not a problem *per se*, since the decoder can decide to attend to any position in the encoder states as long as the state at that position holds the information the decoder needs, this becomes a real issue when we need to make use of the alignment information, as in unknown word replacement (Luong et al., 2015b). As we can see in Figure 2, because of the alignment shift, both **tied** and **fixnorm** incorrectly replace the two unknown words (in bold) with *But Deutsche* instead of *Deutsche Telekom*. In contrast, under **fixnorm+lex** and the model of Arthur et al. (2016), the alignment is corrected, causing the UNKs to be replaced with the correct source words.

### 6.4 Impact of $r$

The single most important hyper-parameter in our models is $r$. Informally speaking, $r$ controls how much surface area we have on the hypersphere to allocate to word embeddings. To better understand its impact, we look at the training perplexity and dev BLEUs during training with different values of $r$. Table 6 shows the train perplexity and best tokenized dev BLEU on Turkish-English for **fixnorm** and **fixnorm+lex** with different values of $r$. As we can see, a smaller $r$ results in

|  | untied | tied | fixnorm | fixnorm+lex | Moses | Arthur |
|---|---|---|---|---|---|---|
| ta-en | 10.3 | 11.1 | 14 (+2.9) | 15.3 (+4.2) | 10.5 (−0.6) | 14.1 (+3.0) |
| ur-en | 7.9 | 10.7 | 12 (+1.3) | 13 (+2.3) | 14.6 (+3.9) | 12.5 (+1.8) |
| ha-en | 16.0 | 16.6 | 20 (+3.4) | 21.5 (+4.9) | 22.2 (+5.6) | 18.7 (+2.1) |
| tu-en | 12.2 | 12.6 | 16.4 (+3.8) | 19.1 (+6.5) | 18.1 (+5.5) | 16.3 (+3.7) |
| uz-en | 14.9 | 15.7 | 18.2 (+2.5) | 19.3 (+3.6) | 17.2 (+1.5) | 17.1 (+1.4) |
| hu-en | 21.6 | 23.0 | 24.0 (+1.0) | 25.3 (+2.3) | 21.3 (−1.7) | 22.7 (−0.3)† |
| en-vi | 25.1 | 25.3 | 26.8 (+1.5) | 27 (+1.7) | 26.7 (+1.4) | 26.2 (+0.9) |
| en-ja (BTEC) | 51.2 | 53.7 | 52.9 (-0.8)† | 51.3 (−2.6)† | 46.8 (−6.9) | 52.4 (−1.3)† |
| en-ja (KFTT) | 24.1 | 24.5 | 26.1 (+1.6) | 26.2 (+1.7) | 21.7 (−2.8) | — |

Table 3: Test BLEU of all models. Differences shown in parentheses are relative to **tied**, with a dagger (†) indicating an *insignificant* difference in BLEU ($p > 0.01$). While the method of Arthur et al. (2016) does not always help, **fixnorm** and **fixnorm+lex** consistently achieve significant improvements over **tied** ($p < 0.01$) except for English-Japanese (BTEC). Our models also outperform the method of Arthur et al. on all tasks and outperform Moses on all tasks but Urdu-English and Hausa-English.

| | | |
|---|---|---|
| input | Dushanba kuni Hindistonda kamida **34** kishi halok bo'lgani xabar qilindi . | |
| reference | At least **34** more deaths were reported Monday in India . | |
| **untied** | At least UNK people have died in India on Monday . | |
| **tied** | It was reported that at least **700** people died in Monday . | |
| **fixnorm** | At least **34** people died in India on Monday . | |
| **fixnorm+lex** | At least **34** people have died in India on Monday . | |
| input | Yarın **Kenya'da** bir yardım konferansı düzenlenecek . | |
| reference | Tomorrow a conference for aid will be conducted in **Kenya** . | |
| **untied** | Tomorrow there will be an **Afghan** relief conference . | |
| **tied** | Tomorrow there will be a relief conference in **Myanmar** . | |
| **fixnorm** | Tomorrow it will be a aid conference in **Kenya** . | |
| **fixnorm+lex** | Tomorrow there will be a relief conference in **Kenya** . | |
| input | Ammo muammolar hali ko'p , deydi amerikalik olim **Entoni Fauchi** . | |
| reference | But still there are many problems , says American scientist **Anthony Fauci** . | |
| **untied** | But there is still a lot of problems , says **James Chan** . | |
| **tied** | However , there is still a lot of problems , says American scientists . | |
| **fixnorm** | But there is still a lot of problems , says American scientist UNK UNK . | |
| **fixnorm+lex** | But there are still problems , says American scientist **Anthony Fauci** . | |

Table 4: Example translations, in which **untied** and **tied** generate incorrect, but often semantically related, words, but **fixnorm** and/or **fixnorm+lex** generate the correct ones.

| | | |
|---|---|---|
| hu-en | 244 | 244 (0.599) document (0.005) By (0.003) by (0.002) offices (0.001) |
| | befektetéseinek | investments (0.151) investment (0.017) Investments (0.015) all (0.012) investing (0.003) |
| | kutatás-fejlesztésre | research (0.227) Research (0.040) Development (0.014) researchers (0.008) development (0.007) |
| tu-en | ifade | expression (0.109) expressed (0.061) express (0.056) speech (0.024) expresses (0.020) |
| | cumhurbaşkanı | President (0.573) president (0.030) Republic (0.027) Vice (0.010) Abdullah (0.008) |
| | Göstericiler | protesters (0.115) demonstrators (0.050) Protesters (0.033) UNK (0.004) police (0.003) |
| ha-en | 😏 | 😏 (0.469) cholera (0.003) EOS (0.001) UNK (0.001) It (0.001) |
| | Wayoyin | phones (0.414) wires (0.097) mobile (0.088) cellular (0.064) cell (0.061) |
| | manzonsa | Prophet (0.080) His (0.041) Messenger (0.015) prophet (0.010) his (0.009) |

Table 5: Top five translations for some entries of the lexical tables extracted from **fixnorm+lex**. Probabilities are shown in parentheses.

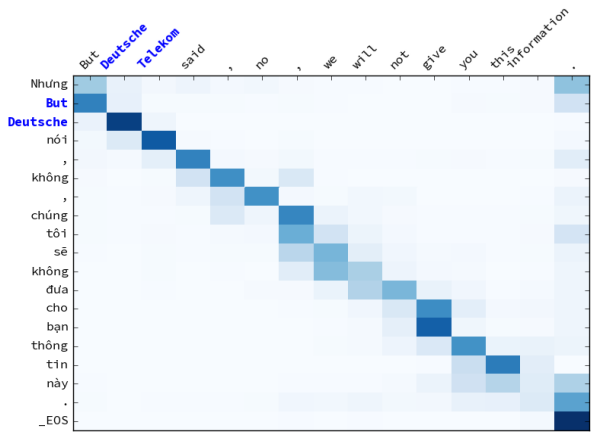

(a) **tied**

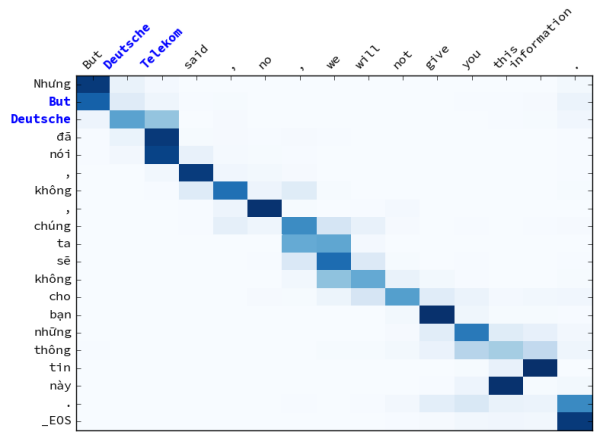

(b) **fixnorm**

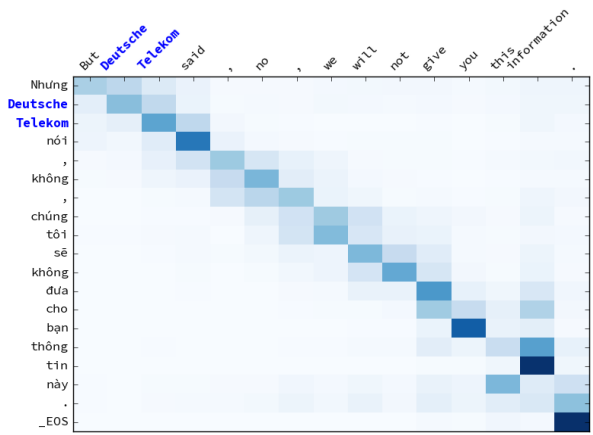

(c) **fixnorm+lex**

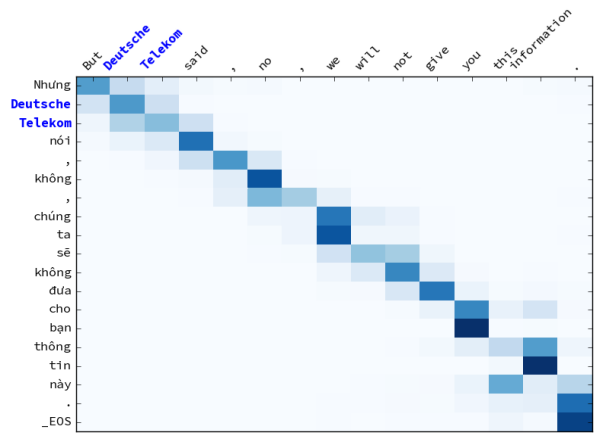

(d) Arthur et al. (2016)

Figure 2: While the **tied** and **fixnorm** systems shift attention to the left one word (on the source side), our **fixnorm+lex** model and that of Arthur et al. (2016) put it back to the correct position, improving unknown-word replacement for the words *Deutsche Telekom*. Columns are source (English) words and rows are target (Vietnamese) words. Bolded words are unknown.

| system | r | train ppl | dev BLEU |
|---|---|---|---|
| **fixnorm** | 3 | 3.9 | 13.6 |
| | 5 | 2.5 | 16.1 |
| | 7 | 2.3 | 14.4 |
| **fixnorm+lex** | 2 | 4.2 | 12.3 |
| | 3.5 | 2.0 | 17.5 |
| | 5 | 1.4 | 16.0 |

Table 6: When $r$ is too small, high train perplexity and low dev BLEU indicate underfitting; when $r$ is too large, low train perplexity and low dev BLEU indicate overfitting.

worse training perplexity, indicating underfitting, whereas if $r$ is too large, the model achieves better training perplexity but decrased dev BLEU, indicating overfitting.

### 6.5 Lexicon

One byproduct of **lex** is the lexicon, which we can extract and examine simply by feeding each source word embedding to the FFNN module and calculating $p^\ell(y) = \text{softmax}(W^\ell h^\ell + b_\ell)$. In Table 5, we show the top translations for some entries in the lexicons extracted from **fixnorm+lex** for Hungarian, Turkish, and Hausa-English. As expected, the lexical distribution is sparse, with a few top translations accounting for the most probability mass.

### 6.6 Byte Pair Encoding

Byte-Pair-Encoding (BPE) (Sennrich et al., 2016) is commonly used in NMT to break words into word-pieces, improving the translation of rare words. For this reason, we reran our experiments using BPE on the LORELEI and English-Vietnamese datasets. Additionally, to see if our proposed methods work in high-resource scenarios, we run on the WMT 2014 English-German (en-de) dataset,[3] using *newstest2013* as the development set and reporting tokenized, case-sensitive BLEU on *newstest2014* and *newstest2015*.

We validate across different numbers of BPE operations; specifically, we try {1k, 2k, 3k} merge operations for ta-en and ur-en due to their small sizes, {10k, 12k, 15k} for the other LORELEI datasets and en-vi, and 32k for en-de. Using BPE results in much smaller vocabulary sizes, so we do not apply a vocabulary cut-off. Instead, we train on an additional copy of the training data in which all types that appear once are replaced with UNK, and halve the number of epochs accordingly. Our models, training, and evaluation processes are largely the same, except that for en-de, we use a 4-layer decoder and 4-layer bidirectional encoder (2 layers for each direction).

Table 7 shows that our proposed methods also significantly improve the translation when used with BPE, for both high and low resource language pairs. With BPE, we are only behind Moses on Urdu-English.

## 7 Related Work

The closest work to our **lex** model is that of Arthur et al. (2016), which we have discussed already in Section 4. Recent work by Liu et al. (2016) has very similar motivation to that of our **fixnorm** model. They reformulate the output layer in terms of directions and magnitudes, as we do here. Whereas we have focused on the magnitudes, they focus on the directions, modifying the loss function to try to learn a classifier that separates the classes' directions with something like a margin. Wang et al. (2017a) also make the same observation that we do for the **fixnorm** model, but for the task of face verification.

Handling rare words is an important problem for NMT that has been approached in various ways. Some have focused on reducing the number of UNKs by enabling NMT to learn from a larger vocabulary (Jean et al., 2015; Mi et al., 2016); others have focused on replacing UNKs by copying source words (Gulcehre et al., 2016; Gu et al., 2016; Luong et al., 2015b). However, these methods only help with unknown words, not rare words. An approach that addresses both unknown and rare words is to use subword-level information (Sennrich et al., 2016; Chung et al., 2016; Luong and Manning, 2016). Our approach is different in that we try to identify and address the root of the rare word problem. We expect that our models would benefit from more advanced UNK-replacement or subword-level techniques as well.

Recently, Liu and Kirchhoff (2018) have shown that their baseline NMT system with BPE already outperforms Moses for low-resource translation. However, in their work, they use the Transformer network (Vaswani et al., 2017), which is quite different from our baseline model. It would be in-teresting to see if our methods benefit the Transformer network and other models as well.

---
[3] https://nlp.stanford.edu/projects/nmt/

|              | tied | fixnorm    | fixnorm+lex |
|--------------|------|------------|-------------|
| ta-en        | 13   | 15 (+2.0)  | 15.9 (+2.9) |
| ur-en        | 10.5 | 12.3 (+1.8)| 13.7 (+3.2) |
| ha-en        | 18   | 21.7 (+3.7)| 22.3 (+4.3) |
| tu-en        | 19.3 | 21 (+1.7)  | 22.2 (+2.9) |
| uz-en        | 18.9 | 19.8 (+0.9)| 21 (+2.1)   |
| hu-en        | 25.8 | 27.2 (+1.4)| 27.9 (+2.1) |
| en-vi        | 26.3 | 27.3 (+1.0)| 27.5 (+1.2) |
| en-de (newstest2014) | 19.7 | 22.2 (+2.5) | 20.4 (+0.7) |
| en-de (newstest2015) | 22.5 | 25 (+2.5)   | 23.2 (+0.7) |

Table 7: Test BLEU for all BPE-based systems. Our models significantly improve over the baseline ($p < 0.01$) for both high and low resource when using BPE.

## 8 Conclusion

In this paper, we have presented two simple yet effective changes to the output layer of a NMT model. Both of these changes improve translation quality substantially on low-resource language pairs. In many of the language pairs we tested, the baseline NMT system performs poorly relative to phrase-based translation, but our system surpasses it (when both are trained on the same data). We conclude that NMT, equipped with the methods demonstrated here, is a more viable choice for low-resource translation than before, and are optimistic that NMT's repertoire will continue to grow.

## Acknowledgements

This research was supported in part by University of Southern California subcontract 67108176 under DARPA contract HR0011-15-C-0115. Nguyen was supported in part by a fellowship from the Vietnam Education Foundation. We would like to express our great appreciation to Sharon Hu for letting us use her group's GPU cluster (supported by NSF award 1629914), and to NVIDIA corporation for the donation of a Titan X GPU. We also thank Tomer Levinboim for insightful discussions.